\documentclass[lettersize,journal]{IEEEtran}
\usepackage{amsmath,amsfonts}
\usepackage{mathrsfs}
\usepackage{bbm}
\usepackage[ruled,vlined]{algorithm2e}
\usepackage{booktabs}
\usepackage{xcolor}
\usepackage{array}
\usepackage[caption=false,font=normalsize,labelfont=sf,textfont=sf]{subfig}
\usepackage{textcomp}
\usepackage{stfloats}
\usepackage{url}
\usepackage{verbatim}
\usepackage{graphicx}
\usepackage{cite}
\begin{document}

\title{Self-evolving Autoencoder Embedded Q-Network}

\author{J. Senthilnath \IEEEmembership{Senior~Member,~IEEE}, Bangjian Zhou, Zhen Wei Ng, Deeksha Aggarwal, Rajdeep Dutta, Ji Wei Yoon, Aye Phyu Phyu Aung, Keyu Wu, Min Wu  \IEEEmembership{Senior~Member,~IEEE}, Xiaoli Li \IEEEmembership{Fellow,~IEEE}
\thanks{This work was supported by the Accelerated Materials Development for Manufacturing Program at the Agency for Science, Technology and Research (A*STAR) via the AME Programmatic Fund by the Agency for Science, Technology and Research under Grant A1898b0043.}
\thanks{J. Senthilnath, Bangjian Zhou, Zhen Wei Ng, Rajdeep Dutta, Ji Wei Yoon, Aye Phyu Phyu Aung, Keyu Wu, and Min Wu are with the Institute for Infocomm Research, Agency for Science, Technology and Research (A*STAR), Singapore 138632 (e-mails: J\_Senthilnath@i2r.a-star.edu.sg, BZHOU006@e.ntu.edu.sg, c170128@e.ntu.edu.sg, rajdeep\_dutta@i2r.a-star.edu.sg, yoon\_ji\_wei@i2r.a-star.edu.sg, aye\_phyu\_phyu\_aung@i2r.a-star.edu.sg, wu\_keyu@i2r.a-star.edu.sg, wumin@i2r.a-star.edu.sg).}
\thanks{Deeksha Aggarwal is with the Center for Data Sciences, International Institute of Information Technology Bangalore (IIITB), India 560100 (e-mail: deeksha.aggarwal003@iiitb.ac.in).}
\thanks{Xiaoli Li is with the Institute for Infocomm Research, Agency for Science, Technology and Research (A*STAR), Singapore, School of Computer Science and Engineering, Nanyang Technological University, Singapore and A*STAR Centre for Frontier AI Research, Singapore 138632 (email: xlli@i2r.a-star.edu.sg).}
\thanks{\noindent This work has been submitted to the IEEE for possible publication. Copyright may be transferred without notice, after which this version may no longer be accessible.}
}

\markboth{Preprint version}%
{Shell \MakeLowercase{\textit{et al.}}: A Sample Article Using IEEEtran.cls for IEEE Journals}


\maketitle

\begin{abstract}
In the realm of sequential decision-making tasks, the exploration capability of a reinforcement learning (RL) agent is paramount for achieving high rewards through interactions with the environment. To enhance this crucial ability, we propose SAQN, a novel approach wherein a self-evolving autoencoder (SA) is embedded with a Q-Network (QN). In SAQN, the self-evolving autoencoder architecture adapts and evolves as the agent explores the environment. This evolution enables the autoencoder to capture a diverse range of raw observations and represent them effectively in its latent space. By leveraging the disentangled states extracted from the encoder generated latent space, the QN is trained to determine optimal actions that improve rewards. During the evolution of the autoencoder architecture, a bias-variance regulatory strategy is employed to elicit the optimal response from the RL agent. This strategy involves two key components: (i) fostering the growth of nodes to retain previously acquired knowledge, ensuring a rich representation of the environment, and (ii) pruning the least contributing nodes to maintain a more manageable and tractable latent space. Extensive experimental evaluations conducted on three distinct benchmark environments and a real-world molecular environment demonstrate that the proposed SAQN significantly outperforms state-of-the-art counterparts. The results highlight the effectiveness of the self-evolving autoencoder and its collaboration with the Q-Network in tackling sequential decision-making tasks.
\end{abstract}


\graphicspath{{Figures/}}

\section{Introduction}
\IEEEPARstart{R}{einforcement} Learning (RL) is a powerful machine learning technique that has gained prominence due to its ability to solve complex research and engineering problems involving sequential decision-making under uncertainty. By enabling Artificial Intelligence (AI) systems to adapt their decision-making processes to evolving  
information and pursue optimal goals, RL has emerged as a practical and impactful approach across 
numerous domains. Its 
potential in real-world applications is evident in areas such as drug design \cite{popova2018deep}, addressing uncertainties in large social networks \cite{aung2022planning}, autonomous driving \cite{zhang2023adaptive}, strategic games \cite{zhu2022empirical} and large-scale classification \cite{wang2021hierarchical}. These examples demonstrate how RL can significantly enhance decision-making capabilities across diverse domains, leading to improved outcomes and advancements.

While significant progress has been achieved in addressing numerous Markov Decision Processes (MDPs), persistent challenges remain. 
One major challenge arises when dealing with large and diverse state and action spaces, whether they are discrete, continuous, or hybrid in nature \cite{whitehead1991complexity}. The computational complexity of finding solutions increases substantially under such circumstances, making the solution procedure computationally intractable. Another challenge arises when an agent is deployed in real-world scenarios lacking 
precise kinematic models. In an ideal setting, the agent would possess 
model providing the state-transition rules and rewards, enabling it to simultaneously learn from real-world data streams and derive an optimal decision-making policy \cite{arulkumaran2017deep}. However, in the absence of such a model, the agent must rely on extensive training to effectively learn the associated model and dynamics. This necessitates a delicate equilibrium 
between exploration where the agent must venture into unexplored or inadequately explored action spaces, 
and exploitation that entails revisiting actions yielding superior rewards. 

In recent years, various techniques, such as restricted Boltzmann machines \cite{hinton2006reducing} and stacked autoencoders (AEs) with encoder-decoder networks \cite{bengio2006greedy,vincent2008extracting}, have been employed to generate \textit{discrete} latent representations in RL. Contrastingly, 
Variational Autoencoders (VAEs) \cite{Kingma2014VAE} and Generative Adversarial Networks (GANs) \cite{goodfellow2014generative} have been employed to generate \textit{continuous} latent representations in RL. These approaches have demonstrated their ability to generate discrete or continuous latent representations, which can be beneficial for RL training. Previous studies have shown that utilizing \textit{fixed} encoder architectures to generate latent space representations can expedite the training process of RL agents \cite{lange2010deep,gregor2019shaping,mohamed2019model}. Building upon this foundation, \textit{our study introduces a novel self-evolving AE approach for RL}. Unlike fixed encoder architectures, our proposed method adaptively self-initializes the network architecture by autonomously adding or pruning hidden neurons. This self-evolving AE dynamically adjusts its structure to capture the most pertinent features for the given RL task. 
Subsequently, the generated latent states from the self-evolving AE are  seamlessly integrated into a 
Q-Network, enabling the RL agent to obtain improved rewards.

By incorporating the self-evolving AE into the RL framework, our approach offers a unique advantage of adaptive network architecture initialization. This empowers 
the RL agent to automatically learn and utilize latent representations that are most conducive to the task's objectives. The self-evolving AE method represents a significant departure from traditional fixed architectures, opening new avenues for enhancing RL training and elevating performance.
In summary, the key contributions of our paper are four-fold: 
\begin{enumerate}
\item During the pre-training phase, our proposed AE network evolves over time and effectively captures a wide range of observations obtained through random exploration of the environment. This evolution yields a compact representation in the latent space, facilitating efficient storage and utilization of diverse observations.
\item By leveraging the pre-trained encoder of the AE, the latent states are seamlessly integrated into the subsequent Q-Network (QN). This integration significantly reduces the number of training episodes required in RL, enabling faster convergence and more efficient learning.
\item We provide a detailed theoretical analysis supporting the bias-variance regulatory strategy employed to evolve the AE architecture. This strategy ensures that the evolving AE captures concise representations of various raw observations in the latent space, laying a robust foundation for the effectiveness of our approach. 
\item 
We conduct comprehensive experiments in three benchmark environments and a real-world molecular environment. The experimental results demonstrate the superiority of our proposed Self-evolving Autoencoder embedded Q-Network (\textbf{SAQN}) in comparison to state-of-the-art counterparts in most cases. 
\end{enumerate}

The rest of the paper is organized as follows: Sections II and III present 
related works and preliminaries, respectively. Section IV introduces 
the SAQN with theoretical insights. Section V conducts an experimental analysis across four environments. 
Conclusions are drawn in Section VI. Further derivation to support theoretical insights of SAQN are presented 
in Appendix A and Appendix B. Lastly, Appendix C provides the environment description. 

\section{Related Works}\label{Literature}
In this section, we provide a concise overview of existing research that combines generative models with RL, highlighting their pertinence to our study.

\subsection{Generative RL}
RL poses two major challenges: (a) deriving an optimal policy when the state-transition function is unknown to the agent, and (b) defining rewards that effectively guide learning. 
An RL agent aims to learn a policy that maximizes the expected total reward while navigating its environment. Previous studies have explored the combination of generative networks, specifically those generating discrete latent representations (such as Autoencoders), with RL approaches tailored for 
discrete state-action spaces (such as Q-Networks). Notably, researchers \cite{lange2010deep,gregor2019shaping,mohamed2019model} have reported that 
incorporating Autoencoders can significantly expedite the training of RL agents.

Combining a generative model with RL can also help mitigate the overestimation errors in Q-values \cite{Lan2019ql}, thereby leading to improved rewards. 
Studies have showcased that embedding Autoencoders within Q-Networks contributes to improved reward acquisition across a spectrum of real-world tasks \cite{DaqnKimura,Savaridassan2022intAEQN}. Additionally, in several domains, pre-training Autoencoders has outperformed 
Principal Component Analysis (PCA) \cite{erhan2009difficulty}, owing to the capacity 
of Autoencoder's encoder-decoder networks to learn compact, nonlinear manifolds \cite{Senthil2024rbm} within the observation space, where distinct features remain uncorrelated in a nonlinear manner.

\subsection{AE Architecture}
In an Autoencoder (AE), an evolving architecture possesses the capability to initiate the generative learning process without an initial structure and exhibits flexibility in handling changing data streams by dynamically adding or discarding hidden neurons \cite{ashfahani2020devdan}. Compared to a fixed-architecture AE, an evolving-architecture AE offers greater adaptability for the following reasons. Firstly, during the learning process, an evolving AE can grow or prune neurons in response to underfitting or overfitting issues, thereby enhancing the model's generalization ability \cite{goutay2019deep}. In contrast, addressing underfitting and overfitting problems in a conventional fixed-architecture AE often necessitates tuning hyperparameters, which can be a tedious and challenging process \cite{Senthil2024rbm}. Secondly, a fixed AE requires separate and proper tuning of its network architecture for different data streams prior to training \cite{wei2023lstm}. Conversely, an evolving AE eliminates the need for such upfront tuning, as its adaptive structure can dynamically adjust itself during the learning process in different environments. Therefore, \textit{this work focuses on an evolving-architecture AE due to its ability to autonomously adapt in diverse environments throughout the learning process}.

\section{Preliminaries}\label{Prelim}
In this section, we provide a mathematical background to explain the preliminary works of our SAQN approach.

\subsection{Problem Definition}

In this study, we consider a Markov Decision Process (MDP) type of control problem. At each timestamp $t$, an RL agent observes its current state $s_t$ and chooses an action $a_t$, which takes it to the subsequent state $s_{t+1}$. Through this state transition, 
the agent receives a reward $r_t$ from the environment \cite{watkins1992q}. In general, an MDP is characterized by a tuple: $\{$a set $S$ of states, a set $A$ of actions, a state transition function $p_a(s,s')=Pr(s'=s_{t+1}|s=s_t, a=a_t)$, a reward function $r_t(s,a)$$\}$. The primary objective 
is to maximize the expected total return, i.e., $\mathbb{E}[R_t|s_t, a_t]$, where $R_t = \sum_{k=0}^{\infty}\gamma^{k} r_{t+k}$ represents the accumulated reward at timestamp $t$ with $\gamma$ representing 
the discount factor ($0 < \gamma < 1$).

\subsection{Q-Network} 

The current research addresses dynamic problems characterized by large state spaces and finite action spaces (discrete), necessitating the representation of the Q-function using a parameterized function approximator such as a neural network (NN). Therefore, a Q-Network (QN) is employed to estimate the action-value function, $\hat{Q}(s, a;~\theta) = \mathbb{E}[R_t|s_t = s, a_t = a]$, where $\theta$ represents its parameters (weights). The action-value function is learned over evolving parameters $\theta_t$ by utilizing batch-wise samples of finite transitions stored in a replay memory, $\{ s_t, a_t, r_t, s_{t+1} \}_{t\geq 0}$, aiming to minimize the squared temporal difference (TD) error \cite{adam2011experience} as follows:
\begin{equation}\label{eqn:DQNloss}
    L_t(\theta_t)=\mathbb{E}_{(s_t, a_t, r_t, s_{t+1}) \sim \mathcal{B}}[(y_t^{QN}-\hat{Q}(s_t,a_t ;~\theta_t))^2]~,
\end{equation}
where ($y_t^{QN}-\hat{Q}$) represents the TD error, and $y_t^{QN}=r_t(s_t, a_t)+\gamma \max_a \hat{Q}(s_{t+1}, a; \theta^-)$ denotes the target to be achieved by the QN at the $t^{th}$ iteration of training, following 
the Bellman optimality criteria \cite{mnih2016asynchronous}. Note that the target network parameters $\theta^-$ are copied from the learning network, i.e. $\theta^- \leftarrow \theta_t$, only at a regular intervals and are held constant between updates. 
In practice, we choose an $\epsilon$-greedy policy, wherein 
the greedy action, $a = argmax_a \hat{Q}(s,a)$, is chosen 
with probability $1-\epsilon$, while 
a random action is selected with probability $\epsilon$. Additionally, 
an experience replay memory $\mathcal{B}$ is leveraged to improve the batch-wise optimization process for a stable learning \cite{adam2011experience,wei2021deep}.

\subsection{Estimation error in Q-values}  
Q-Network induces a positive bias by approximating the expected action value with the maximum action value 
\cite{AvgDQN_2017,Thrun_1993}. At iteration $i$, let the value function estimated by the QN be 
$\hat{Q}(s,a|\theta_t)$, whereas the true value is $Q^*(s,a)$. The total estimation error at iteration $t$ can be decomposed as \cite{AvgDQN_2017}: $ \hat{Q}(s,a|\theta_t) - Q^*(s,a) = \delta_t = \{ \hat{Q}(s,a|\theta_t) -  y_{s,a}^t \} + \{ y_{s,a}^t - \hat{y}_{s,a}^t \} + \{ \hat{y}_{s,a}^t- Q^*(s,a) \}$, where $y_{s,a}^t=\mathbb{E}_{\mathcal{B}}[r_t + \gamma \max\limits_{a'}Q_{\theta_{t-1}}(s',a'|s,a)]$ is the QN target and $\hat{y}_{s,a}^t = \mathbb{E}_{\mathcal{B}}[r_t + \gamma \max\limits_{a'}(y_{s',a'}^{t-1}|s,a) ]$ denotes the true target \cite{AvgDQN_2017}.

The three components in the total error $\delta_t$ represent the \textit{Target Approximation Error} (TAE), the \textit{Overestimation Error} (OE), and the \textit{Optimality Difference}, respectively. The TAE emerges from factors such as: 
(a) inexact parameter optimization, (b) limited representation power of a NN, and (c) finite size of a buffer memory. The OE introduces a positive bias, causing sub-optimal policy generation, which is non-uniform and gets bigger with states where Q-values are similar for different actions. Essentially, 
a high variance of the TAE in turn increases the OE \cite{Thrun_1993,AvgDQN_2017,ExpRepl_2020}. The OE can be reduced by disentangling the input data correlations. Recall that a latent space representation is compact, wherein different features exhibit much lower correlations 
than in a raw observation space. Therefore, 
mitigating estimation errors in Q-values can be achieved by transmitting latent states instead of raw states, motivating the fusion of an AE with a QN.
\begin{figure*}[ht]
\vskip 0.2in
\begin{center}
\centerline{\includegraphics[height=4.5cm, width=16.5cm]{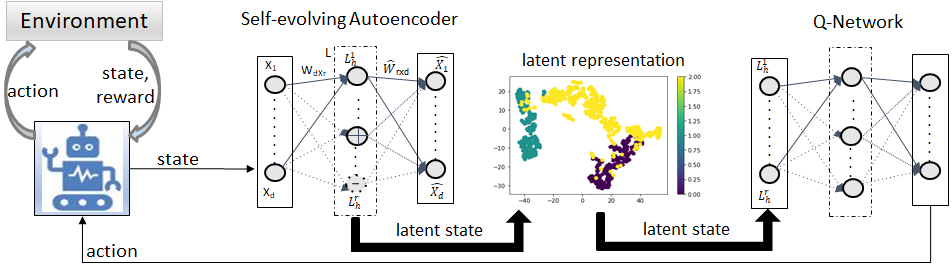}}
\caption{Schematic diagram of the SAQN [+ indicates adding neuron; - indicates pruning neuron]}
\label{fig1}
\end{center}
\vskip -0.2in
\end{figure*}

\subsection{Generating latent states}
The latent distributions of the raw states can be obtained by three means: AEs, VAEs and GANs. When combined with Q-Network, continuous latent representations from VAE or GAN will yield continuous state and action spaces for the underlying RL problem. Consequently, the $\epsilon$-greedy policy, where the Q-Network identifies the greedy action $a = argmax_a \hat{Q}(s,a)$ will become problematic, as it has to find the global maximum for an infinite continuous space at every step~\cite{silver2014deterministic}. Although CAQL \cite{ryu2019caql} proposes a solution for maximizing Q-values with continuous actions using mixed integer programming, it remains computationally expensive for RL frameworks, especially for problems with originally discrete action space. Another approach involving VAE or GAN with Q-Networks utilizes approximation functions (linear or NN), but this can lead to divergence~\cite{sutton2018reinforcement}. Hence, for dynamic problems with discrete state and action spaces, we consider AE which generates discrete latent representations of the raw states of a problem ~\cite{hinton2006reducing, bengio2006greedy, vincent2008extracting} to combine with Q-Network.

\section{Our Proposed SAQN Methodology} \label{SAQN_method}
The proposed self-evolving autoencoder embedded Q-Network (SAQN) method operates on 
two levels: (i) a self-evolving autoencoder  extracts a concise latent representation from the raw observations via evolving neurons; (ii) a Q-Network (QN) evaluates the quality of state-action combinations for optimal decision making. The schematic representation  
of our SAQN methodology is illustrated in Figure \ref{fig1}.

\subsection{Self-evolving Autoencoder}
A self-evolving autoencoder embedded QN is composed of an autoencoder (AE) based on the self-evolving generator ($\mathcal{G}$). During the pre-training phase, the AE learns to create a non-linear manifold of the raw space observed by an explorative agent. It generates a latent space representation where various observations are compressed in an uncorrelated fashion. For an input $X \in \Re^{1 \times d}$, the AE is used to reconstruct the output $\Hat{X} \in \Re^{1 \times d}$ through the encoding-decoding scheme. 

By using the \textit{tanh} activation function, the encoder and decoder outputs are expressed as:   
\begin{equation}
      L = g_t(XW + a); \Hat{X} = g_t(LW^T + b),
\label{Eq. 2}
\end{equation}
where $W \in \Re^{d\times r}$ is the weight matrix; $a \in \Re^{1 \times r}$ and $b \in \Re^{1 \times d}$ are the hidden layer $L$'s and the output layer $\Hat{X}$'s bias vectors, respectively, with $r$ being the number of hidden neurons evolved using a bias-variance regulatory strategy; and $g_t$ denotes the \textit{tanh} activation function.
During the pre-training phase, the hidden layer's neurons are added and pruned based on the bias-variance regulatory strategy \cite{ashfahani2020devdan}. In a self-evolving hidden layer, a hidden neuron ($L_h^i$) where $i \in \{1,...,r\}$ is added or pruned to cope with an under-fitting condition due to the high bias or an over-fitting condition due to the high variance, respectively. The benefit is to learn the input observation space efficiently by updating the network with the features from the recently explored space. 

Note that the expectation of the squared reconstruction error ($\mathbb{E}[e^2]$) between the desired ($X$) and the predicted ($\hat{X}$) outputs can be expressed as:
\begin{align}
\mathbb{E}[e^2] & = \mathbb{E}[(X-\hat{X})^2] = \mathbb{E}[\hat{X}^2] - \mathbb{E}[\hat{X}]^2 + (\mathbb{E}[\hat{X}]-X)^2 \notag\\
& = var(\hat{X}) + (bias(\hat{X}))^2
\label{Eq. 3}
\end{align}
where, $(bias(\hat{X}))^2$ is the bias and $var(\hat{X})$ is the variance of the reconstructed output $\hat{X}$. 
The detailed derivation is included in the Appendix \ref{sec:BIAS and VARIANCE}.

\subsection{Analysis of bias-variance in Self-evolving Autoencoder} 
To derive the expected squared reconstruction error, $\mathbb{E}[e^2] = \mathbb{E}[(X-\Hat{X})^2]$, $(bias(\hat{X}))^2$ and $var(\hat{X})$ need to be evaluated. Consequently, in  (\ref{Eq. 3}), the two terms, $\mathbb{E}[\hat{X}]$ and $\mathbb{E}[\hat{X}^2]$, need to be calculated. Using (\ref{Eq. 2}), the expectations of $L$ and $\hat{X}$ are obtained as
\begin{align}
\label{Eq. 5}
\mathbb{E}[L] & = \mathbb{E}[g_t(F)]  = \int_{-\infty}^{\infty} P(F)g_t(F)\, dF~; \\
\mathbb{E}[\Hat{X}] & = g_t(\mathbb{E}[L]W^T+b)~,
\label{HS}
\end{align}
where $F = XW + b$ and $\ P(F) = \frac{1}{\sqrt{2\times\pi(\sigma^t)^2}}e^{-\frac{(F-\mu^t)^2}{2(\sigma^t)^2}}$ is the normal probability density of $F$ with mean $\mu^t$ and variance $\sigma^t$,  considering observations up to time $t$.
Note that the \textit{tanh} ($g_t$) and the \textit{sigmoid} ($g_s$) activation functions are related by: $g_t(F)=2g_s(2F)-1$, since $F \sim \mathscr{N}(\mu,\sigma^2)$, then $2F \sim \mathscr{N}(2\mu,4\sigma^2)$. By leveraging this relation in conjunction with the probit approximation \cite{murphy2012machine}, (\ref{Eq. 5}) is expressed as
\begin{align}
    \mathbb{E}[L]\approx 2g_s (\frac{2\mu^t}{\sqrt{1+\pi(\sigma^t)^2/2}})-1~, \label{Eq. 6}
\end{align}

\noindent Next, using (\ref{HS}), $E[\hat{X}^2]$ is derived as
\begin{align}
\mathbb{E}[\hat{X}^2] & = g_t(\mathbb{E}[L^2]W^T+b).
\label{Eq. 7}
\end{align}
Recall that the product of two independent identically drawn variables gives $L^2 = L*L$, which leads to:
\begin{align}
\mathbb{E}[\hat{X}^2]  & = g_t(\mathbb{E}[L]^2W^T+b).
\label{Eq. 10}
\end{align}
Then, $\mathbb{E}[\hat{X}]$ is calculated by substituting (\ref{Eq. 6}) into (\ref{HS}), and $\mathbb{E}[\hat{X}^2]$ is calculated by substituting (\ref{Eq. 6}) into (\ref{Eq. 10}), which are given by 
{\footnotesize
\begin{align}
\label{Eq. 17}
\mathbb{E}[\hat{X}] & \approx g_t((2*g_s(\frac{2\mu^t}{\sqrt{1+\pi(\sigma^t)^2/2}})-1)W^T+b)\\
\mathbb{E}[\hat{X}^2] & \approx g_t ((2*g_s(\frac{2\mu^t}{\sqrt{1+\pi(\sigma^t)^2/2}})-1)^2W^T+b)
\label{Eq. 18}
\end{align}
}%
Further, $(bias(\hat{X}))^2$ and $var(\hat{X})$ can be evaluated by substituting $\mathbb{E}[\hat{X}]$ and $\mathbb{E}[\hat{X}^2]$ into (\ref{Eq. 3}).
An equivalent derivation considering the sigmoid activation function, is included in the Appendix \ref{sigmoid deriv}. To alleviate high bias and high variance in the reconstructed output, the hidden neurons in $L$ are evolved by monitoring the following conditions. 
\begin{align}
\mu_{bias}^t + \sigma_{bias}^t \geq \mu_{bias}^{min} + d_1\sigma_{bias}^{min} \label{Eq. 7}\\
\label{Eq. 8}
\mu_{var}^t + \sigma_{var}^t \geq \mu_{var}^{min} + 2d_2\sigma_{var}^{min}
\end{align}
where $\mu_{bias}^t$, $\sigma_{bias}^t$ and $\mu_{var}^t$, $\sigma_{var}^t$ are the recursive mean and standard deviation of the bias and variance (including the current observation), respectively. $\mu_{bias}^{min}$, $\sigma_{bias}^{min}$ and $\mu_{var}^{min}$, $\sigma_{var}^{min}$ are the minimum values of the mean and standard deviation of the bias and variance for all the observations encountered during training but are reset every time the condition (\ref{Eq. 7}) or (\ref{Eq. 8}) is satisfied and a node is added or pruned. Furthermore, $d_1$ is a dynamic constant that is calculated as, $d_1 = \alpha e^{(-(bias(\hat{X}))^2)}+\beta$ in (\ref{Eq. 7}), where $\alpha$ and $\beta$ are constants and the bias is the current observation (bias condition for all the environments: $\alpha$ =1.3 and $\beta$=1.0). In (\ref{Eq. 8}), $d_2$ is another dynamic constant calculated as, $d_2 = \alpha e^{(-(var(\hat{X}))^2)}+\beta$, and the multiplier $2$ is to avoid an immediate pruning just after growing (variance condition for all the environments: $\alpha$ =1.3 and $\beta$=0.7).

In the presence of high bias and high variance in $\hat{X}$, if (\ref{Eq. 7}) is satisfied then the growing of a neuron with Xavier initialized weights is triggered, or if (\ref{Eq. 8}) is satisfied then the pruning of a neuron contributing least to the network output $\hat{X}$ is triggered. The least contributing neuron ($i$) with the minimum $\mathbb{E}[L]$ is determined using (\ref{Eq. 6}) as $prune: \min_{i=1,\dots,n} \mathbb{E}[L]_i$. This notion of the statistical contribution for growing or pruning neurons is the backbone of a bias-variance regulatory strategy for a self-evolving AE to induce generalization capability into the network's output estimates.

Followed by the pre-training of the self-evolving AE, a QN, i.e. $\mathcal{Q}$, is trained to predict the optimal actions. The hidden layer $L$ consists of the latent representation where different features are captured in an uncorrelated way to be passed as the states into the following $\mathcal{Q}$. Overall, the proposed SAQN is summarized in \textbf{Algorithm 1} and \textbf{Algorithm 2}.

\begin{algorithm}
  \KwIn{Observation replay memory \textbf{X} containing raw states $\{O_t\}_t$ $\in { \rm I\!R}^d$; \textbf{$O_t$} is generated by random agent movements following a Gaussian policy.}
  \KwOut{Reconstructed output $\mathbf{\Hat{X}}$ to samples \textbf{X}, Latent space of dimension \textbf{r}, Pre-trained encoder layer L}\par
  \textbf{Pre-train AE}:\par
  \While{$t\leq maxSteps$}{
    Sample \textbf{$O_t$} batch-wise from \textbf{X} \par
    \For{x in \textbf{$O_t$}}{
    \textbf{Calculate:} \text{ $\mathbb{E}[\Hat{x}]$, $\mathbb{E}[L]$ and $\mathbb{E}[\Hat{x}^2]$} \text{using  [\ref{HS}], [\ref{Eq. 6}], [\ref{Eq. 10}]}\par
    \text{$\mu^{t}_{bias}, \sigma^{t}_{bias},\mu^{t}_{var}, \sigma^{t}_{var}$} \text{using [\ref{Eq. 7}], [\ref{Eq. 8}]}\;
        
    \textbf{Hidden Node Growing:}\par
    \eIf{($\mu^{t}_{bias} + \sigma^{t}_{bias} \geq \mu^{min}_{bias} + d_1\sigma^{min}_{bias}$)}{ 
        \textbf{Initialization:} \par\text{Using xavier-uniform to generate new} \par\text{row of weight and bias}\par
        \textbf{Reset:} \textbf{$\mu^{min}_{bias},\sigma^{min}_{bias}$}\par
        \text{r $\gets$ r + 1}\par
        \text{$Flag_{grow}$ $\gets$ 1}
    }{
        \text{$Flag_{grow}$ $\gets$ 0}
    }
    \textbf{Hidden Node Pruning:}\par
    \If{($\mu^{t}_{var} + \sigma^{t}_{var} \geq \mu^{min}_{var} + 2d_2\sigma^{min}_{var}$) \text{and} ($Flag_{grow}$ != 1) \text{and} (r $>$ 1)}{
        \textbf{Delete Rows and Cols}:\par\text{Prune node with least $\mathbb{E}[L]$}\par
        \textbf{Reset:} \textbf{$\mu^{min}_{var},\sigma^{min}_{var}$}\par
        \text{r $\gets$ r - 1}\par
    }
    Compute reconstruction loss between $\mathbf{\Hat{X}}$ and \textbf{X}
  }
  \text{t $\gets$ t + 1}\par
  }
  \caption{Pre-train self-evolving Autoencoder}
\end{algorithm}

\begin{algorithm}
    \KwIn{Representation replay memory $\mathcal{B}$ containing latent states $\{s_t\}_t$ $\in { \rm I\!R}^r$, the evolved encoder layer \textbf{L} from \textbf{Algorithm 1}, Concatenate pre-trained encoder layer \textbf{L}'s output in the QN input layer.}
    \KwOut{$Q^*, a^*$ ($*$ denotes the optimal)}
    \textbf{Train with QN}:
    \par iteration: t $\gets$ 0, train$\_$interval = $\Delta$ t
    \par \While{environment is not done}{
    Sample $s_t$ batch-wise from $\mathcal{B}$ \par
    Calculate policy: $\pi(s_t) \leftarrow argmax_a Q(s_t,a_t)$\par
    Determine action: $a_t \leftarrow$ $\pi(s_t)$\par
    State transition: $s_{t+1} \leftarrow s_t, a_t$ \par
    Obtain reward: $r_t(s_t, a_t)$ \par
    Store information $<s_t, a_t, r_t, s_{t+1}>$ into $\mathcal{B}$\par    
    \If{t$\%$ $\Delta$ t == 0}
    {
    Sample a minibatch of transitions from $\mathcal{B}$\par
    Find $\max_{a_{t+1}} Q(s_{t+1},a_{t+1};~\theta)$ \par
    Update future action value $Q(s_{t+1},a_{t+1})$\par
    Minimize loss (\ref{eqn:DQNloss}) by training QN($\theta$)
    }\par
    Update current action value $Q(s_t,a_t)$\par
    \text{t $\gets$ t + 1}\par
 }
 \caption{Train SAQN}
\end{algorithm}
\begin{table*}[h!]
\centering
\caption{Average performance metrics and standard deviation (over 10 or 100 trials) in each environment}
\label{tab:Table2}
\begin{tabular}{lllll}
\toprule
{\color[HTML]{000000} Environments} & 
{\color[HTML]{000000} Metrics} & 
{\color[HTML]{000000} QN} &
{\color[HTML]{000000} AQN} & 
{\color[HTML]{000000} SAQN} \\ 
\cmidrule(r){1-5}

{\color[HTML]{000000} CartPole-v0} & 
{\color[HTML]{000000} \begin{tabular}[c]{@{}r@{}}NCS\\ AEC\\ ATT \end{tabular}} & 
{\color[HTML]{000000} \begin{tabular}[c]{@{}l@{}} 76/100 \\ 212.27±20.53 \\ \textbf{60.67}±10.13  \end{tabular}} & 
{\color[HTML]{000000} \begin{tabular}[c]{@{}l@{}} 84/100 \\ 165.26±22.95\\ 64.80±4.95 \end{tabular}} & {\color[HTML]{000000} \begin{tabular}[c]{@{}l@{}} \textbf{87/100}\\ \textbf{140.26}±33.26\\ 82.63±7.49 \end{tabular}} \\ 
\cmidrule(r){1-5}

{\color[HTML]{000000} LunarLander-v2} & 
{\color[HTML]{000000} \begin{tabular}[c]{@{}r@{}}NCS\\ AEC\\ ATT \end{tabular}} & 
{\color[HTML]{000000} \begin{tabular}[c]{@{}l@{}}59/100\\548.71±87.22\\1267.44±250.07 \end{tabular}} & 
{\color[HTML]{000000} \begin{tabular}[c]{@{}l@{}}96/100\\390.82±82.30 \\815.67±136.83\end{tabular}} &
{\color[HTML]{000000} \begin{tabular}[c]{@{}l@{}}\textbf{100/100}\\\textbf{276.89}±76.00\\\textbf{794.52}±113.25\end{tabular}} \\ 
\cmidrule(r){1-5}

{\color[HTML]{000000} Minigrid} & 
{\color[HTML]{000000} \begin{tabular}[c]{@{}r@{}}AR\\ ATT \end{tabular}} & 
{\color[HTML]{000000} \begin{tabular}[c]{@{}l@{}}-15.80±3.85\\224.52±52.76\end{tabular}} & 
{\color[HTML]{000000} \begin{tabular}[c]{@{}l@{}}-10.12±2.46\\219.11±57.01\end{tabular}} &
{\color[HTML]{000000} \begin{tabular}[c]{@{}l@{}}\textbf{-9.93}±3.24\\\textbf{218.78}±67.55\end{tabular}} \\
\cmidrule(r){1-5}

{\color[HTML]{000000} Molecular optimization} & 
{\color[HTML]{000000} \begin{tabular}[c]{@{}r@{}}AR\\ATT \end{tabular}} & 
{\color[HTML]{000000} \begin{tabular}[c]{@{}l@{}}0.6058±0.0067\\\textbf{17229.25}±526.59\end{tabular}} & 
{\color[HTML]{000000} \begin{tabular}[c]{@{}l@{}}0.6280±0.0054\\ 44054.24±2110.10\end{tabular}} &
{\color[HTML]{000000} \begin{tabular}[c]{@{}l@{}}
\textbf{0.6618}±0.0125\\42357.32±1672.38\end{tabular}} \\ 
\bottomrule
\end{tabular}%
\end{table*}
\section{Experiments}
Comprehensive experiments have been conducted to evaluate the performance of the proposed SAQN in comparison to the state-of-the-art methods across four environments. 

\subsection{Experimental settings}
The performance of the proposed SAQN is compared with two state-of-the-art techniques: i) Q-Network \cite{Mnih2013DRLs} and ii) Integrated autoencoder and Q-Network (AQN) \cite{DaqnKimura, Savaridassan2022intAEQN}. These RL techniques are applied to three benchmark problems from OpenAI Gym (CartPole-v0, LunarLander-v2, Minigrid) and a real-world molecular environment. The detailed descriptions of all four environments are included in the Appendix section \ref{sec:env desc}. Each simulation is run for either 10 (for complex environment) or 100 (for simple environment) times, and the average results are tabulated. These simulations are run on an i7-10875H CPU processor at 2.30GHz with 16GB RAM without GPU acceleration. 

\noindent\textbf{Implementation details.} To make a fair comparison, the Q-Network for all three techniques has the same structure, i.e., input is passed into a hidden layer of 256 neurons, before passing to the output layer, which gives out the Q-values of each action. As our proposed SAQN evolves neurons in the single layer, hence, fixed AQN method of pre-training with autoencoder is set to a single layer.
\begin{table}
\centering
\caption{The number of neurons and activation functions}
\label{encoder structure}
\begin{tabular}{lllll}
\hline
Environments & Input & AQN/SAQN & Act. in QN & Act. in AE\\
\hline
CartPole-v0  & 4 & 12  &  tanh & tanh  \\
LunarLander-v2 & 8 & 23 & tanh & tanh\\
Minigrid   & 147 & 268 & sigmoid & sigmoid \\
Molecular  & 2049 & 2052 & tanh & relu \\
\hline
\end{tabular}
\end{table}

\begin{figure}[ht]
    \centering   \includegraphics[height=4cm,width=8.1cm]{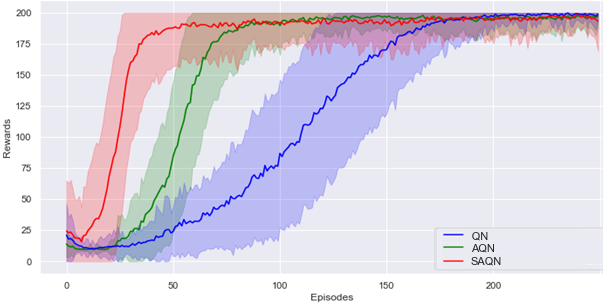}
   \caption{\centering Average reward vs. episodes by different RL agents in the Cartpole environment.}
    \label{Figure:Cart-reward-episodes}
\end{figure}
\begin{figure}[ht]
    \centering   \includegraphics[height=4cm,width=8.1cm]{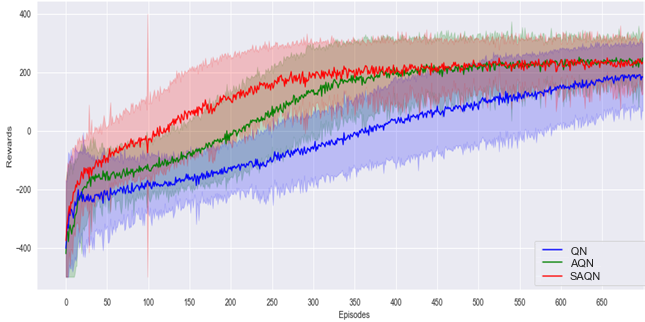}
   \caption{\centering RL agent average reward vs. episodes in the Lunarlander-v2.}
    \label{Figure:Lun-reward-episodes}
\end{figure}
\begin{figure}[ht]
    \centering   \includegraphics[height=4cm,width=8.1cm]{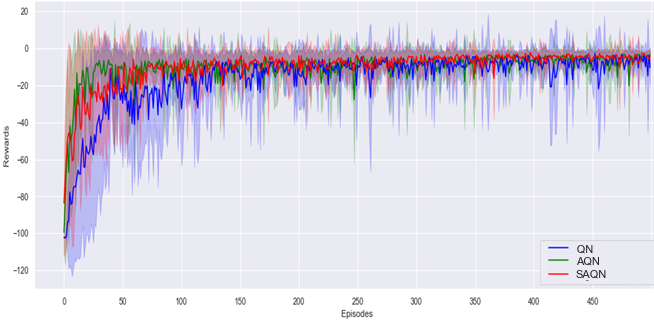}
   \caption{\centering Average reward vs. episodes by different RL agents in the Minigrid environment.}
    \label{Figure:Mini-reward-episodes}
\end{figure}
\begin{figure}
    \centering
    [a] \includegraphics[width=0.43\linewidth]{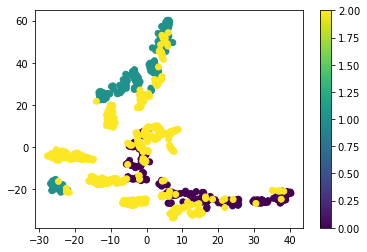}
    [b] \includegraphics[width=0.43\linewidth]{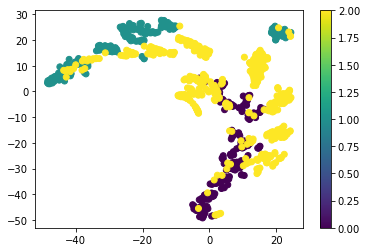}
    [c] \includegraphics[width=0.43\linewidth]{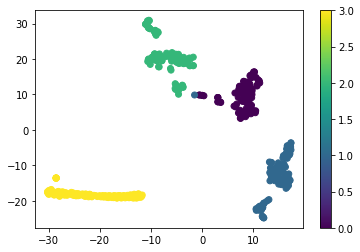}
    [d] \includegraphics[width=0.43\linewidth]{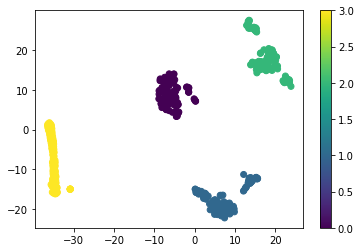}
   \caption{\centering TSNE plots of raw observation states for [a] Cartpole-v0 [c] LunarLander-v2; latent representation of pre-trained SAQN encoder layer for [b] Cartpole-V0; [d] LunarLander-v2.}
    \label{Figure:cartpole-lun-latent}
\end{figure}
\noindent\textbf{Quantitative Evaluation.}
We first give some common definitions of the symbols and characters used here: $N_s$ is the number of simulations, $N_e$ is the number of episodes in each simulation, $Boolean$ is the boolean function-if the related condition is satisfied it would get 1 otherwise 0, $thr$ is the corresponding stopping threshold. Each technique is evaluated using four metrics as follows: i) Number of Converged Simulations (NCS), ii) Average Episodes to Convergence (AEC), iii) Average Time Taken per simulation (ATT), and iv) Average Reward per simulation (AR).
\vspace{-0.3cm}
\begin{equation}
  NCS = \sum_{i=1}^{N_s} Boolean (r_{\{i,N_e\}}\ge thr)
  \label{Mul1}
\end{equation}
\vspace{-0.8cm}
\begin{multline}
  AEC = (\sum_{i=1}^{N_s} \min_{x\in A_i} x)/N_s,A_i= \{x|1\leq x \leq N_e\quad\\ \&\& \quad r_{\{i,x\}}\ge thr\}
\end{multline}
\vspace{-0.4cm}
\begin{equation}
   ATT = (\sum_{i=1}^{N_s} \sum_{x=1}^{N_e} time_{\{i,x\}})/N_s
\end{equation}
\begin{equation}
   AR = (\sum_{i=1}^{N_s} \sum_{x=1}^{N_e} reward_{\{i,x\}})/N_s
 \end{equation}
The average reward in the last 100 episodes starting from the $N_e^{th}$ episode of simulation $i$ is denoted by $r_{\{i,N_e\}}$; the average reward in the last 100 episodes starting from the $x^{th}$ episode of simulation $i$ is denoted by $r_{\{i,x\}}$; $time_{\{i,x\}}$ is the time consumed during the $x^{th}$ episode of simulation $i$ in seconds; and $reward_{\{i,x\}}$ is the final reward in the $x^{th}$ episode of simulation $i$. Note that the AR metric is used when there is no specific stopping threshold.
\begin{figure}
    \centering
    [a] \includegraphics[width=0.43\linewidth]{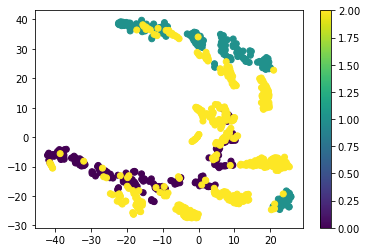}
    [b] \includegraphics[width=0.43\linewidth]{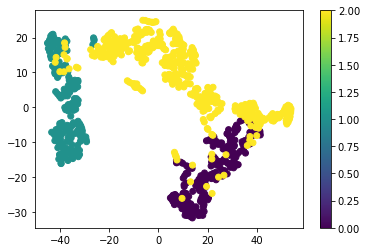}
   \caption{\centering TSNE plots of Cartpole-v0 [a] output from regular QN’s fully-trained 1st layer; [b] latent space representation of fully-trained SAQN encoder layer.}
    \label{Figure:cartpole-latent}
\end{figure}
\begin{figure*}
    \centering   \includegraphics[height=4.52cm,width=16.1cm]{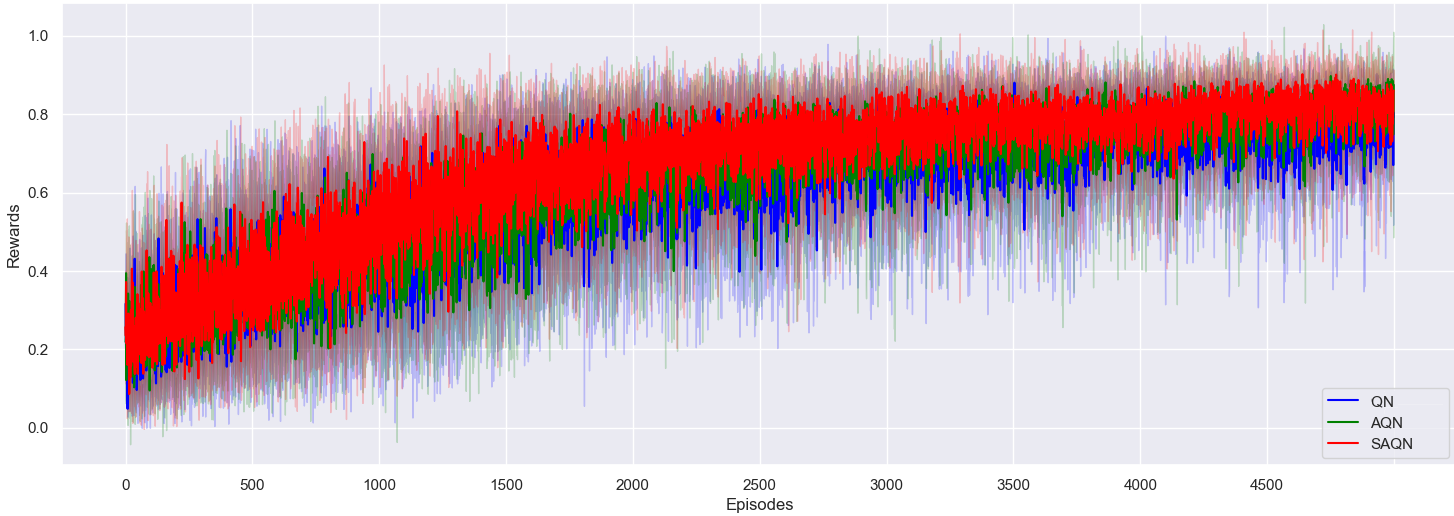}
   \caption{\centering Average reward vs. episodes by different RL agents in the Molecular environment.}
    \label{Figure:Mol-reward-episodes}
\end{figure*}
\subsection{Evaluation of benchmark environment}
\label{sec:benchmark}
The performance evaluation of three different RL methods on three benchmark environments and a real-world environment is shown in Table \ref{tab:Table2}. 

In CartPole-v0 and LunarLander-v2 environments, the proposed SAQN outperforms the other two RL methods, QN and AQN, in terms of the stopping threshold indicating a convergence criteria. A higher NCS indicates that the SAQN performance is more stable under random weight and random policy initializations, and a higher AEC justifies the quick learning ability of the SAQN agents rendering fast convergence in these two environments. In CartPole-v0, the performance improvement of SAQN over QN comes at the price of computational time consumption, as indicated by the corresponding ATT values. The ATT value with the SAQN is almost 1.4 times higher than that of the QN because the environment is simple and SAQN has more parameters in its encoder layer. In LunarLander-v2, as the dimension of latent states increases, the QN agent usually takes around 200 steps in each episode and its ATT is 1.6 times higher than that of the proposed SAQN method. As a result, the QN enables firing the main engine more often and the lander drops slowly whereas the SAQN enables a quick landing. Figure \ref{Figure:Cart-reward-episodes} shows the average reward performance at every episode. The reward trend shows that in CartPole-v0, SAQN is the fastest one providing less variations initially but more at the end. The final rewards achieved with these three RL methods are close as the underlying environment is less complex. In Lunarlander-v2 (Figure \ref{Figure:Lun-reward-episodes}), the AQN is better than the QN throughout the learning process, which justifies the benefit of pre-training AE. Further, the SAQN learns faster than the AQN due to its evolving AE network.

For the benchmark environment Minigrid, the AR metric is used to evaluate as this environment don't have a specific stopping criteria. In Minigrid, the SAQN method performs the best and slightly better than the AQN. From Figure \ref{Figure:Mini-reward-episodes}, it is clear that the SAQN performs better than the QN with less variations though these variations are slightly more than the AQN. Since the proposed SAQN can meet the objective faster in each episode, even with twice parameters as compared to the QN, the SAQN still has a lower ATT.

\begin{table}
\centering
\caption{Molecular QED distribution}
\begin{tabular}{llll}
\hline
QED & QN & AQN & SAQN \\
\hline
$\geq$ 0.90  & 302 & 826 & \textbf{1430} \\
$\geq$ 0.85 & 1325 & 2708 & \textbf{4109} \\
$\geq$ 0.80 & 4103 & 5907 & \textbf{7782} \\
\hline
\end{tabular}
\label{molecular_distribution}
\end{table}
\noindent\textbf{Latent space visualization.} After training the RL agent in the Cartpole-v0 environment, the associated self-evolving encoder generated latent space is visualized in Figure \ref{Figure:cartpole-lun-latent} [a] and [b], which depicts three categories of states in the memory that have been used to pre-train the self-evolving AE structure: i) pole angle $>$ $-15^{\circ}$ (purple dots), ii) pole angle $>$ $+15^{\circ}$ (green dots) and iii) pole angle within the scope $\pm15^{\circ}$ and distance to the center between $\pm2.4$ units (yellow dots). Note that `-' is turn to left and `+' is turn to right. The termination criteria is triggered by the first two categories, while purple and green dots denote the `bad states' and yellow dots denote the `good states'. The observation state distribution in Figure \ref{Figure:cartpole-lun-latent} [a] show the inter-category overlaps, especially between purple and yellow dots. However, the output of the pre-trained SAQN encoder shown in Figure \ref{Figure:cartpole-lun-latent} [b], reveals yellow dots are compact with some separation from purple dots in the latent representation. In the Lunarlander-v2 environment (Figure \ref{Figure:cartpole-lun-latent} [c] and [d]), four categories of states have been used: i) both legs landed on the ground (purple dots), ii) left leg landed on the ground (blue dots), iii) right leg landed on the ground (green dots) and iv) none of the legs landed on the ground (yellow dots). In the raw observation space shown in Figure \ref{Figure:cartpole-lun-latent} [c], there exists a blue dot in the margin between purple and green dots. However, such overlap is not present in the encoder-generated latent space and different categories are well-separated as shown in Figure \ref{Figure:cartpole-lun-latent} [d]. This justifies the pre-training of an encoder to pass uncorrelated samples for a stable training of the QN. 

Also, in Cartpole-v0 environment, the observation state distribution (Figure \ref{Figure:cartpole-latent} [a]) show the inter-category overlaps still persists with QN's fully-trained state distribution. However, the output of the latent space representation of fully-trained SAQN encoder layer in Figure \ref{Figure:cartpole-latent} [b], reveals that different categories in the latent representation are relatively well-separated than that of the raw observation space (Figure \ref{Figure:cartpole-lun-latent} [a]) and QN's fully-trained state distribution (Figure \ref{Figure:cartpole-latent} [a]). Such uncorrelated states fed from the latent space to the QN justifies its improved performance.

\subsection{Evaluation of molecular optimization}
\label{sec:moldqn1}
The real-world environment, molecular optimization, is complex as the underlying observation space is high-dimensional and sparse. In this complex environment, the evolving AE architecture of SAQN shows more effectiveness due to its capability of pruning and growing hidden neurons dynamically. The AR value obtained with SAQN is 0.6618, which is significantly better than that of AQN (0.6280) and QN (0.6058). This result justifies that the latent space generated by the encoder layer helps extracting compact features that in turn benefits training the QN to improve AR value. Note that the expected number of hidden neurons in the encoder layer is usually less than the dimension of raw observations (2049), though it is slightly higher (2052) in the current case (Please see the Table \ref{encoder structure}).

In Figure \ref{Figure:Mol-reward-episodes}, it is observed that the rewards obtained with SAQN increase to be slightly higher than that of the AQN towards the end of training. Also, the SAQN performance is stable in relatively high QED region at the end of training. However, this performance improvement comes at the price of computational time consumption. Due to the involvement of additional parameters, the SAQN is 2.45 times slower than the regular QN. 

\begin{figure}
    \centering
    \includegraphics[width=0.9\linewidth]{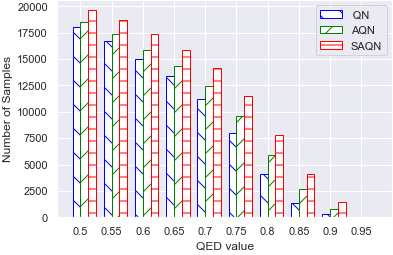}
   \caption{\centering Number of samples over different QED values}
    \label{Figure:molecular-freq}
\end{figure}

The objective of the molecular optimization is to discover high QED structures. The SAQN method has shown an ability to discover novel molecular structures with high QED. The highest QED values obtained with QN, AQN, and SAQN are 0.946, 0.948, and 0.947, respectively. Throughout the training episodes, the SAQN is able to produce molecules with $QED\geq0.90$ in 1430 occasions (Table \ref{molecular_distribution}), whereas the AQN and the QN produced such molecules in 826 and 302 occasions, respectively. Further, the results in Figure \ref{Figure:molecular-freq} demonstrate that SAQN has the ability to uncover many molecular structures with high QED values. Even for the $QED\geq0.80$ or $\geq0.85$, the SAQN is able to produce 1.3 to 3.1 times more structures than the other two RL methods. For $QED\leq0.5$, the three RL methods uncovered similar structures.
From Figure \ref{Figure:molecular-freq}, it is clear that for $0.5<QED<0.75$, the SAQN performs still better than the AQN and QN. We attribute this performance improvement to the evolving AE architecture of SAQN, which promotes the newly generated hidden nodes to adapt to a variety of high-dimensional sparse information and suits well in this case with many different types of molecular structures. It is worth noting that the AQN's one encoder layer with the same dimension as that of the SAQN cannot perform very well, which might be due to its fixed architecture wherein all hidden neurons always participate during the training phase. Overall, the proposed SAQN method is well-performing in all of the tested environments. Note that forty state-transitions are considered in each training episode here, and due to the involvement of more parameters (hidden neurons), the ATT corresponding to SAQN is higher than that of QN but lesser than the AQN. This shows the benefits of evolving architecture (SA) over fixed architecture (AE).
\begin{figure*}[ht!]
    \centering
    [a] \includegraphics[width=0.75\linewidth]{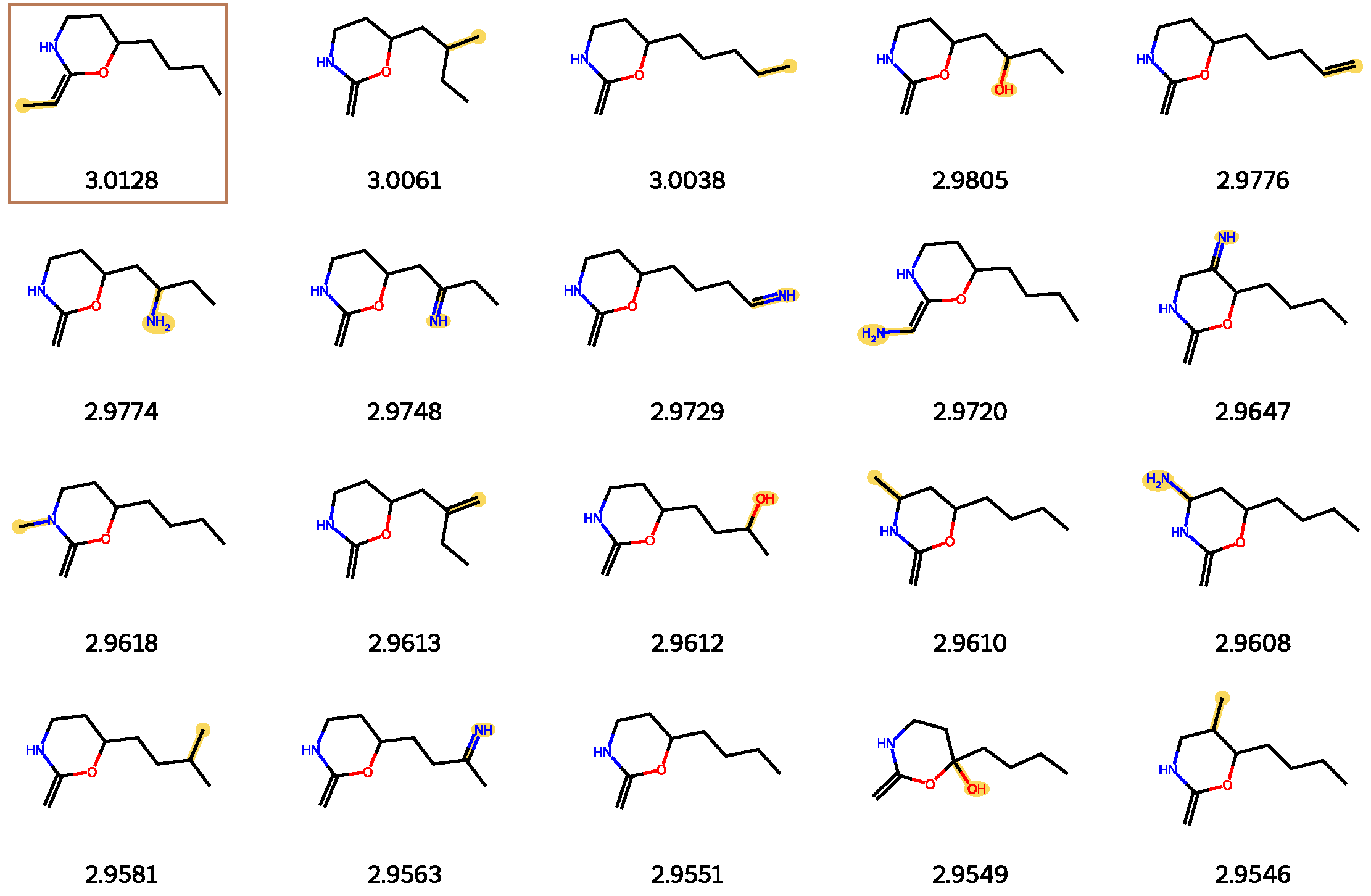}\\
  \vspace{0.1cm}
   \hspace{7in}
    [b] \includegraphics[width=0.75\linewidth]{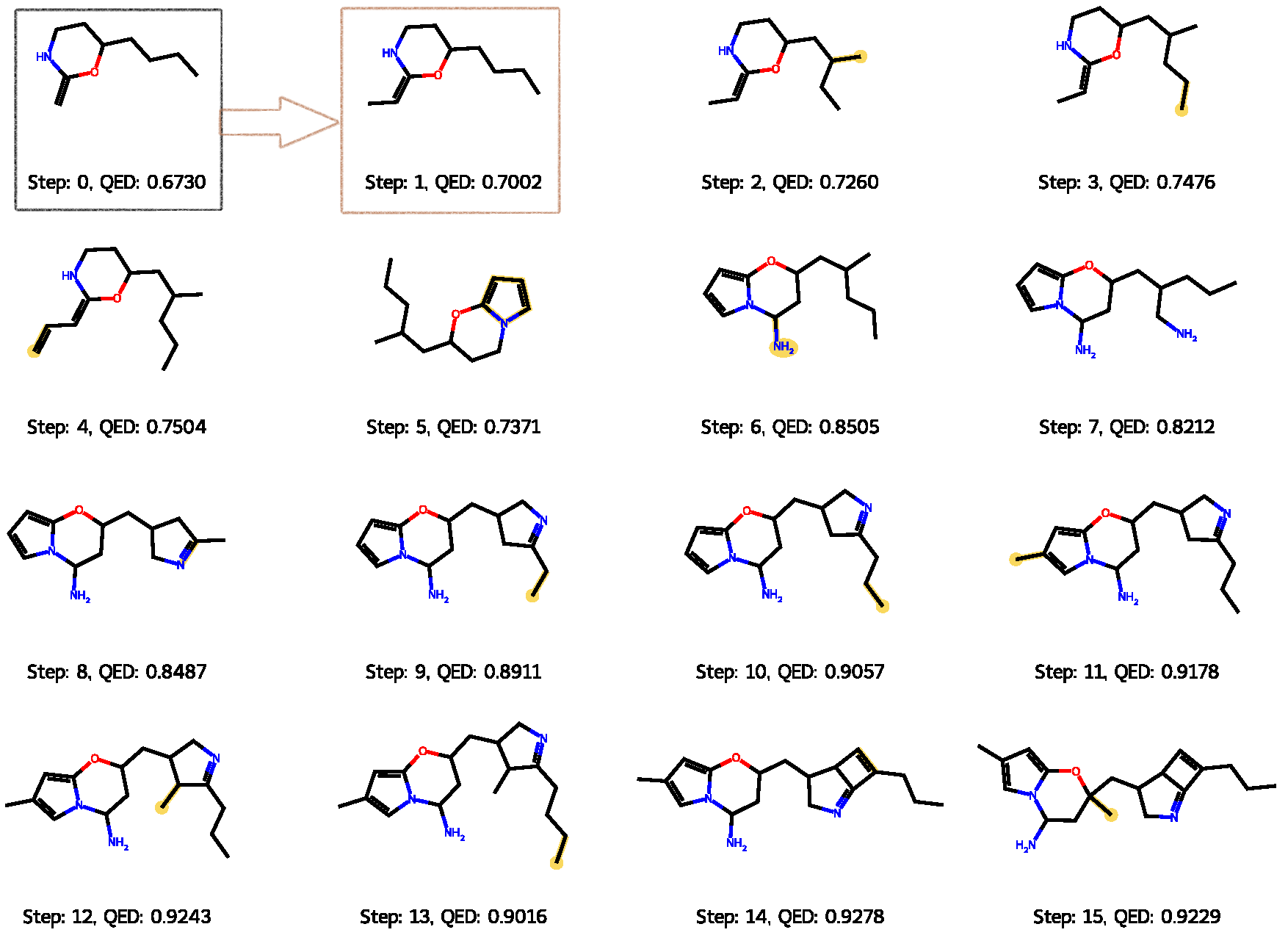}
   \caption{\centering Molecular structure change [a] Set of possible actions for step 0 and its corresponding Q-values, the box indicates maximum Q-value selection; [b] the process by which the changes of the structure for maximum Q-value from Step 0 to Step 1.}
    \label{Figure:molecular-q-values}
\end{figure*}

\noindent\textbf{Visualization of molecular structure evolution.}
Here, we visualize the evolution of molecular structures using SAQN. Similar structures can have drastically different simplified molecular-input line-entry system (SMILES) representations, which challenges an AE to form a compact latent space. Figure \ref{Figure:molecular-q-values} [a] shows the candidate actions at step 0. From this figure, we can observe that the highest Q-value of 3.0128 is achieved by adding a bond at the left part of the structure, while adding a hydroxide at the right part (2.9805) is not favored by the agent. The agent would always take the action that produces the highest Q-value. Although QED has an implicit relationship with Q-value, it might drop slightly during the evolution as shown in Figure \ref{Figure:molecular-q-values} [b] (steps 4 to 5). 
However, throughout the process, we observe that by picking the highest Q-value action, the agent could make the structure evolve to attain a higher QED. The usefulness of the Q-Network has also been discussed in \cite{Zhou2019MOLDQN}.

\section{Conclusion}
In this paper, a self-evolving AE embedded with a QN (SAQN) is proposed to design an effective RL agent. The proposed evolving AE architecture is able to adapt to different observation spaces in a variety of benchmark environments, and shows its impact by improving the obtained rewards with an effective exploration of the search space. Further, the exploration induced by our proposed SAQN method facilitates in optimal decision-making even in a complex molecular environment with a high-dimensional sparse observation space. This study focuses on the environments with discrete action spaces for a clear demonstration of the impact of a latent representation to efficiently train a QN. Given the performance of SAQN, we believe the proposed method can also be applied to other value-based RL methods so that they can improve performance and adapt to different dynamic problems.
This will lead to our future work where we will investigate how to effectively use VAE or GAN to represent the continuous latent states and then, use policy-based RL methods (actor-critic or policy gradient RL approaches) for continuous action spaces \cite{lillicrap2015continuous, mnih2016asynchronous}.



{\appendices
\section{Bias-Variance derivation in SAQN}\label{sec:BIAS and VARIANCE}
\begin{align}
\begin{split}
E[e^2] 
& = E[(\Hat{X}-X)^2] = E[(\Hat{X}-E[\Hat{X}]+E[\Hat{X}]-X)^2]\\
& = E[(\Hat{X}-E[\Hat{X}])^2]+E[2(\Hat{X}-E[\Hat{X}])(E[\Hat{X}]-X)]\\
&\,\,\,\,\,\,+E[(E[\Hat{X}]-X)^2]
\end{split}
\end{align}
Since $\Hat{X}$ and $X$ are i.i.d, we have $E[\Hat{X}*X]=E[\Hat{X}]*E[X]$. The middle term in the above equation is solved as follows:
\begin{align}
\begin{split}
& E[2(\Hat{X}-E[\Hat{X}])(E[\Hat{X}]-X)] \\
& = 2(E[\Hat{X}*E[\Hat{X}]-E[\Hat{X}]*E[\Hat{X}]-\Hat{X}*X+E[\Hat{X}]*X])\\
& = 2(E[\Hat{X}]*E[\Hat{X}]-E[\Hat{X}]*E[\Hat{X}]-E[\Hat{X}]*E[X]\\
&\,\,\,\,\,\,+E[\Hat{X}]*E[X])\\
& = 0
\end{split}
\end{align}
Hence, we obtain $E[e^2]$ as follows:
\begin{align}
\begin{split}
E[e^2]
& = E[(\Hat{X}-E[\Hat{X}])^2]+E[(E[\Hat{X}]-X)^2]\\
& = E[(\Hat{X}^2-\Hat{X}*E[\Hat{X}]-E[\Hat{X}]*\Hat{X}+E[\Hat{X}]^2)]\\
&\,\,\,\,\,\, +E[(E[\Hat{X}]-X)^2]\\
& = E[\Hat{X}^2] - E[\Hat{X}]^2 + (E[\Hat{X}]-X)^2\\
& = var(\Hat{X}) + (bias(\Hat{X}))^2
\end{split}
\end{align}

\section{Derivation for sigmoid activation in SAQN}
\label{sigmoid deriv}
When we apply the sigmoid activation function ($g_s$), the mathematical expressions of $E[\Hat{X}]$ and $E[\Hat{X}^2]$ change as follows.
\begin{align}
E[\Hat{X}] & = g_s(E[L]W^T+c)\label{Eq. 11}
\end{align}
\begin{align}
E[L] & = \int_{-\infty}^{\infty} P(F)g_s(F)\, dF\label{Eq. 12}
\end{align}
Using the probit approximation \cite{murphy2012machine}, ``\eqref{Eq. 12}" is expressed as
\begin{align}
\label{Eq. extanh}
E[L] & \approx g_s(\frac{\mu^t}{\sqrt{1+\pi(\sigma^t)^2/8}})
\end{align}
$E[\hat{X}]$ is obtained by substituting ``\eqref{Eq. extanh}" into ``\eqref{Eq. 11}". From ``\eqref{Eq. 11}", $E[\hat{X}^2]$ can be expressed as
\begin{align} \label{Eq.Ex2}
E[\hat{X}^2] & = g_s(E[L]^2W^T+c),
\end{align}
Consequently, $E[\hat{X}^2]$ is obtained by substituting ``\eqref{Eq. extanh}" into ``\eqref{Eq.Ex2}". 

\section{Environment description}
\label{sec:env desc}
\subsection{CartPole-v0}
The CartPole environment is a game whereby a pole is attached to a cart that moves along a frictionless track. The pole is initialized upright, and the goal of the agent is to prevent it from falling over either the pole angle is more than 15 degrees from the vertical axis, or the cart leaves the centre more than 2.4 units. The agent takes in a 4 dimensional observation states vector $s =(X,{\dot{X},\theta,\dot{\theta}}) $, which contains the horizontal distance with the velocity and pole angle with its angle velocity, and chooses 2 possible actions: pushing left, or pushing right. The agent will win a $+1$ score for each time-step if it keeps the pole upright. The environment is considered "solved" when the agent receives an average reward of 195 over 100 consecutive episodes
\footnote{\url{https://www.gymlibrary.ml/envs/CartPole-v0/}accessed on January 2022}.

\subsection{Lunarlander-v2}
The LunarLander environment is a game whereby a lander (UFO) which has 2 supportive legs tries to land within a landing pad without crashing and firing as little of its engine as possible. The agent takes in 8-dimensional observation states, which include a list of information like xy coordinates, corresponding velocity, angle, angle velocity and flags to indicate whether 2 legs have landed or not. And chooses 4 possible actions: do nothing, fire the left orientation engine, fire the main engine, or fire the right orientation engine. Each time firing the engine will win the agent a $-0.3$ score. And there will be additional plus/minus points considering whether the lander comes to the rest area and the legs conditions in the end. The environment is considered "solved" when the agent receives an average reward of 200 over 100 consecutive episodes \footnote{\url{https://gym.openai.com/envs/LunarLander-v2/} accessed on January 2022}.

\subsection{Minigrid}
The Minigrid environment is a game whereby the agent starts from some point in a room and tries to search and reach the goal.
We loaded the environment $'MiniGrid-Empty-16x16-v0'$ from the gym package, which has 16x16 cells and with no obstacles and no doors inside the room. The agent takes in 147-dimensional observations states that include the current position and the information of its searching area and chooses 7 possible actions: including moving in 4 directions and 3 interactions with objects, which are doors, keys, and walls on the map. 
The reward policy is 
when the agent reaches the goal, it will add a positive score, and for each step it took, it will get a negative score \footnote{\url{https://github.com/maximecb/gym-minigrid}}. The environment also does not gives a 'solved' condition, so we use AR to compare.

\subsection{Molecular optimization}
The molecular environment discussed in \cite{Zhou2019MOLDQN} where an agent suggests the molecular structure. In return, the environment will calculate the corresponding quantitative estimate of drug-likeness (QED) and give out a corresponding reward. It also returns an addition or reduction of the current structure. The agent takes in 2049 dimensional observation states - including 2048 simplified molecular-input line-entry system (SMILES) code to depict the molecular structure and 1 digit to describe how many steps are left in this episode and gives out the index of the action in the list of valid actions in the next step. These valid actions include adding/removing an atom and adding/removing a bond. We set the maximum number of steps for each episode as 40 that is in each episode we try 40 molecular structures. Unlike the above three environments, the objective is to discover some excellent molecular structures \footnote{\url{https://github.com/deepchem/torchchem/tree/master/contrib/MolDQN}}. In this environment, we use the AR metric for comparison purposes.

}


\bibliographystyle{IEEEtran}
\bibliography{SAQNbib}

\vfill

\end{document}